\documentclass[11pt]{article}

\usepackage[final]{acl}

\usepackage{times}
\usepackage{latexsym}

\usepackage[T1]{fontenc}
\usepackage[utf8]{inputenc}

\usepackage{microtype}

\usepackage{lmodern}
\usepackage{graphicx}

\usepackage{makecell} % Required for \makecell command
\usepackage{multirow}
\usepackage{tabularx}
\usepackage{caption}       % optional, for caption tweaks
\usepackage{amsmath}
\usepackage{array}
\usepackage[table]{xcolor}
\usepackage{float}  % in preamble
\usepackage{cuted}

\usepackage{booktabs}

\usepackage{hyphenat} % allows breaking in \texttt without forcing htt font shapes

\usepackage{needspace}           % avoid orphaned section headers
\usepackage{placeins}

\usepackage[most]{tcolorbox}
\tcbset{
  colback=gray!5,
  colframe=black!50,
  boxrule=0.5pt,
  arc=2pt,
  outer arc=2pt,
  left=6pt,
  right=6pt,
  top=6pt,
  bottom=6pt
}

\title{Operationalising the “Right to be Forgotten” in LLMs:\\
A Lightweight Sequential Unlearning Framework for Privacy-Aligned Deployment in Politically Sensitive Environments}

\author{
    Esen Kurt \\
    Department of Mathematics \\
    Munster Technological University \\
    {\tt esen.kurt@mymtu.ie}
    \And
    Haithem Afli \\
    Department of Computer Science \\
    Munster Technological University \\
    {\tt haithem.afli@mtu.ie}
}

\begin{document}
\maketitle

\begin{abstract}

Large Language Models (LLMs) are increasingly deployed in politically sensitive environments, where memorisation of personal data or confidential content raises regulatory concerns under frameworks such as the GDPR and its “right to be forgotten”. Translating such legal principles into large-scale generative systems presents significant technical challenges.

We introduce a lightweight sequential unlearning framework that explicitly separates retention and suppression objectives. The method first stabilises benign capabilities through positive fine-tuning, then applies layer-restricted negative fine-tuning to suppress designated sensitive patterns while preserving general language competence.

Experiments on the SemEval-2025 LLM Unlearning benchmark demonstrate effective behavioural suppression with minimal impact on factual accuracy and fluency. GPT-2 exhibits greater robustness than DistilGPT-2, highlighting the role of model capacity in privacy-aligned adaptation. We position sequential unlearning as a practical and reproducible mechanism for operationalising data erasure requirements in politically deployed LLMs.

\end{abstract}

% ------------------------------------------------------------------
\section{Introduction}
% ------------------------------------------------------------------

\begin{figure}[t]
    \centering
    \includegraphics[width=\linewidth]{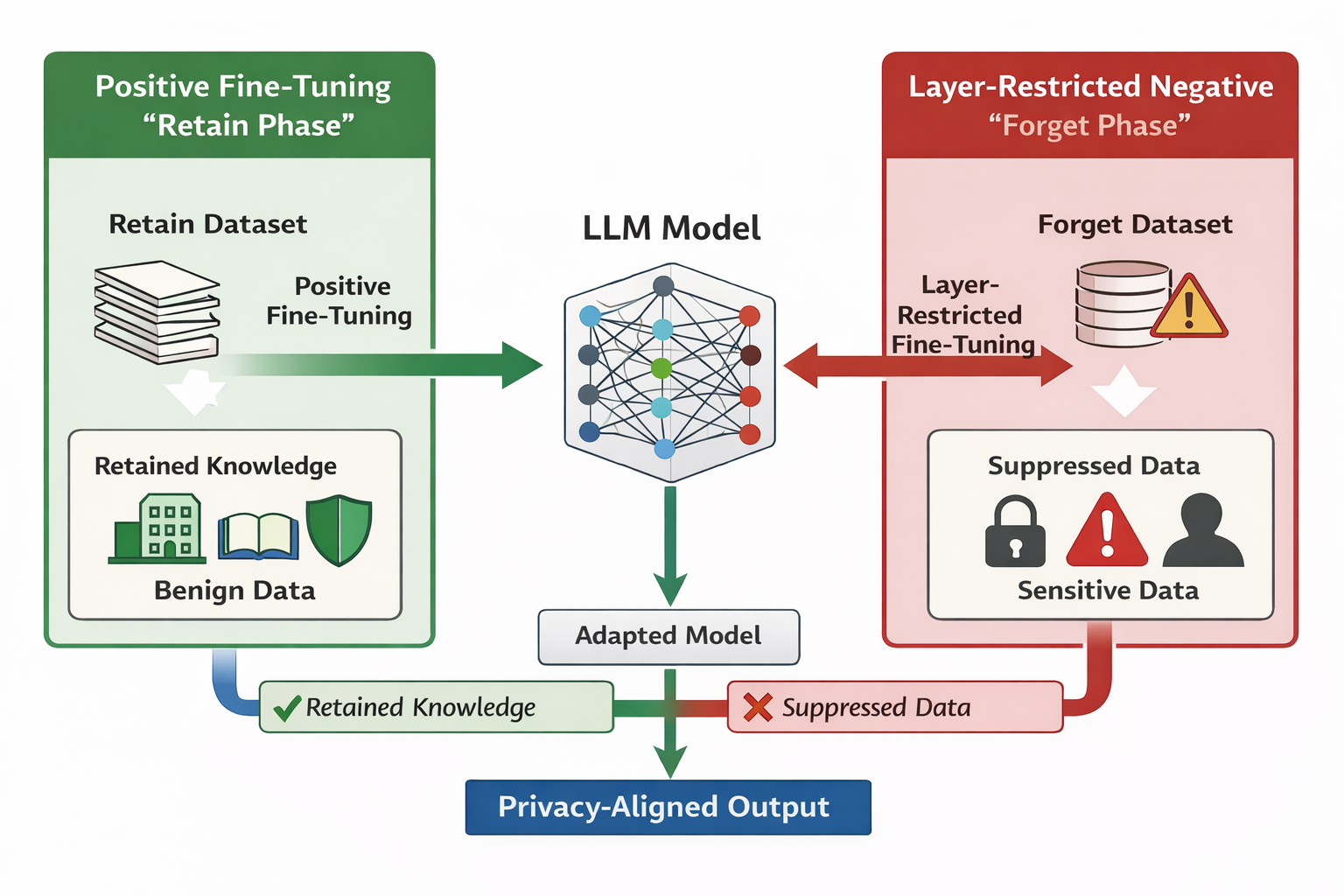}
    \caption{
    Sequential unlearning framework for operationalising the “right to be forgotten” in Large Language Models. 
    The model is first stabilised through positive fine-tuning on a Retain dataset (benign knowledge), 
    followed by layer-restricted negative fine-tuning on a Forget dataset (sensitive patterns). 
    By separating retention and suppression into distinct optimisation phases, the approach reduces gradient interference 
    while preserving general language competence and enabling privacy-aligned outputs.
    }
    \label{fig:sequential_unlearning}
\end{figure}

Large Language Models (LLMs) are increasingly embedded within politically sensitive infrastructures, including electoral communication systems, journalistic tools, public-sector chatbots, and civic information platforms. As generative models become intermediaries of public discourse, their internal representations function as a form of algorithmic memory. However, large-scale pretraining on web corpora can lead to the unintended memorisation and reproduction of sensitive political content, including personal data of public officials, confidential communications, or unverified allegations \citep{carlini2021extracting}. 

Such behaviour raises not only technical safety concerns, but also regulatory and democratic challenges. Under frameworks such as the General Data Protection Regulation (GDPR) and the California Consumer Privacy Act (CCPA), individuals retain the “right to be forgotten”, establishing legal obligations for the removal of personal data from automated systems. Translating this legal principle into the context of large-scale generative models is non-trivial: retraining models from scratch is computationally prohibitive, and full deletion of internal representations remains difficult to verify.

Machine unlearning has therefore emerged as a research direction aimed at reducing or suppressing the influence of specific training data while preserving overall model utility. Existing approaches span full retraining, counterexample-driven fine-tuning, gradient-based loss manipulation, and direct parameter editing. While full retraining offers stronger theoretical guarantees, it is rarely feasible in politically deployed LLM systems where models must be adapted rapidly in response to regulatory or reputational risk. More practical approaches focus on behavioural suppression, modifying model outputs without requiring complete retraining.

In politically sensitive environments, however, suppression must be carefully balanced with retention. Over-aggressive forgetting may degrade factual accuracy, historical accountability, or civic knowledge, while insufficient suppression may expose private or destabilising information. This tension motivates the need for structured optimisation strategies that explicitly separate the objectives of retaining benign knowledge and discouraging sensitive reproduction.

To address these challenges, we propose a sequential unlearning framework that operationalises the “right to be forgotten” in LLMs through a lightweight two-phase optimisation regime, illustrated in Figure~\ref{fig:sequential_unlearning}. Rather than jointly optimising retention and forgetting signals, we stabilise useful capabilities before applying targeted suppression. Experiments conducted using the SemEval-2025 LLM Unlearning Challenge benchmark \cite{ramakrishna2025unlearning} 
provide a controlled evaluation of this approach. Our objective is to reduce the model’s ability to reproduce sensitive patterns while preserving general language competence, thereby aligning privacy protection with democratic information integrity.
% ------------------------------------------------------------------
\subsection{Sequential unlearning: separating retention and forgetting}
% ------------------------------------------------------------------

Many existing unlearning approaches treat retention and forgetting as a single joint optimisation problem, encouraging models to perform well on ``retain'' data while simultaneously degrading performance on ``forget'' examples. This formulation often introduces gradient interference, resulting in unstable convergence, incomplete suppression, or widespread catastrophic forgetting.

We instead propose a \textbf{sequential two-phase optimisation regime}:

\begin{enumerate}
    \item \textbf{Positive Fine-Tuning (Retain phase):} reinforcement of benign, factual, and non-sensitive behaviours on a curated Retain dataset, anchoring the model within a stable region of parameter space.
    
    \item \textbf{Layer-Restricted Negative Fine-Tuning (Forget phase):} targeted gradient ascent on a Forget dataset, applied only to the final transformer blocks and language modelling head to discourage reproduction of sensitive patterns while preserving lower-layer linguistic representations.
\end{enumerate}

By separating retention and suppression into distinct optimisation stages, the framework reduces gradient conflict and enables more interpretable trade-offs between utility and forgetting. Empirically, this leads to more predictable suppression of sensitive content and significantly less collateral degradation of general language ability.

Interestingly, our results also highlight differences in model robustness: GPT-2 remains stable under sequential unlearning, whereas DistilGPT-2 exhibits performance collapse, suggesting that unlearning may serve as a stress test for representational capacity and resilience.

% ------------------------------------------------------------------
\subsection{Framework efficiency and deployment relevance}
% ------------------------------------------------------------------

Although evaluated in a benchmark setting, the proposed framework is designed with real-world political deployment constraints in mind.

\begin{itemize}
    \item It is \textbf{lightweight}: no full retraining or large-scale safety corpus construction is required; suppression is achieved through limited parameter updates and low learning rates.
    
    \item It is \textbf{behaviour-oriented}: the method discourages sensitive template reproduction at higher transformer layers rather than attempting exhaustive identification of memorised content.
    
    \item It is \textbf{architecture-agnostic}: the sequential regime can be adapted to other autoregressive transformer models.
    
    \item It explicitly \textbf{separates privacy alignment from utility preservation}, reducing the risk that politically motivated unlearning degrades factual knowledge or civic usefulness.
\end{itemize}

We therefore position sequential unlearning as a practical mechanism for operationalising data erasure rights in LLMs deployed within politically sensitive contexts. Beyond its technical contribution, this work contributes to broader discussions on AI governance, democratic accountability, and the normative implications of model memory in computational social science \cite{batool2025aigov}.

% ------------------------------------------------------------------
\section{From Web Erasure to Model Memory: The Evolution of the “Right to be Forgotten”}
% ------------------------------------------------------------------

\subsection{Origins in Web Search and Indexing}

The “right to be forgotten” (RTBF) emerged in response to the persistence of
personal data in digital archives and search engine indexing. Early debates
centred on whether individuals should be able to request the removal of
outdated or harmful personal information from search results, even when such
information remained legally published elsewhere. The landmark \emph{Google
Spain v. AEPD and Mario Costeja González}  \citep{google_spain_2014}
decision of the Court of Justice of the European Union established that search engines could be
required to delist certain personal data upon request, thereby recognising
a practical form of digital erasure.

The introduction of Article 17 of the General Data Protection Regulation (GDPR) in 2018 
 \citep{eu_gdpr_2016} formalised this principle within EU law, granting individuals
the right to request erasure of personal data under specified conditions.
Similar, though not identical, provisions appear in the UK GDPR, the
California Consumer Privacy Act (CCPA) and its amendment under the CPRA 
 \citep{ccpa_2018}, as well as in emerging data protection regimes across Latin America and
Asia-Pacific jurisdictions.

\subsection{From indexed content to learned representations}

While early RTBF enforcement focused on web pages and search engine links,
the rise of large-scale machine learning systems complicates the notion of
erasure. Unlike search engines, which index external documents, Large
Language Models (LLMs) internalise statistical patterns from massive
pretraining corpora. Information is no longer stored as discrete retrievable
documents, but as distributed representations embedded within model
parameters \cite{shilov2026mosaic}.

This architectural shift introduces new challenges. In web search,
removal typically involves delisting or deleting specific records. In LLMs,
however, sensitive information may be encoded across millions of parameters
in a non-localised manner. Consequently, the technical meaning of
“erasure” becomes ambiguous: does compliance require full retraining,
parameter-level modification, or behavioural suppression at inference time?

\subsection{RTBF in the era of generative AI regulation}

The rapid deployment of generative AI systems has prompted renewed
regulatory attention. The European Union’s AI Act \citep{eu_ai_act_2024},
alongside evolving interpretations of GDPR enforcement, places emphasis on transparency,
risk mitigation, and governance mechanisms for high-impact AI systems.
In the United States, state-level data protection laws such as the CCPA/CPRA
grant deletion rights, though their applicability to model parameters
remains legally unsettled. Other jurisdictions, including Brazil (LGPD)
and India’s Digital Personal Data Protection Act, similarly recognise
forms of erasure rights  \citep{global_dp_laws_2023}.

Across these regimes, a central question persists: how can data subject
rights be meaningfully exercised when information has been absorbed into
large neural models? Current regulatory texts rarely specify technical
requirements for compliance in generative systems, leaving substantial
interpretative gaps between legal obligations and engineering practice.

\subsection{Operational challenges for LLMs}

Operationalising the “right to be forgotten” in LLMs therefore requires
rethinking the relationship between data, memory, and model behaviour.
Full retraining to exclude specific data points may be infeasible for
large models due to computational and financial costs. Conversely,
purely output-level filtering may not satisfy stronger interpretations
of erasure.

In this sense, unlearning transforms RTBF from a document-level governance tool into a parameter-level intervention within generative infrastructures. By reducing or suppressing the
influence of designated data without retraining from scratch,
unlearning techniques provide a pragmatic approach to compliance.
However, they also raise normative questions: to what extent does
behavioural suppression constitute meaningful erasure? How should
conflicts between privacy rights, public interest, and historical
accountability be resolved?

As LLMs increasingly mediate political communication and civic discourse,
these questions become central to AI governance. The evolution of the
“right to be forgotten” from web indexing to generative model memory
marks a shift from document-level control to parameter-level governance,
requiring new technical frameworks capable of balancing privacy protection,
model utility, and democratic accountability.

% ------------------------------------------------------------------
\section{Related Work}
% ------------------------------------------------------------------

\paragraph{Machine unlearning and data erasure in AI systems.}
Machine unlearning was initially developed in the context of classical machine
learning, where the goal was to efficiently remove the influence of specific
training examples without requiring full retraining
\citep{bourtoule2021machine,ginart2019making}. 
With the rise of Large Language Models (LLMs), unlearning has gained renewed
attention due to the scale of pretraining corpora and the increasing
deployment of generative systems in socially and politically consequential
domains\cite{satvaty2026survey}.

In LLMs, memorisation of sensitive content raises not only technical safety
concerns but also regulatory and governance challenges, particularly under
data protection frameworks such as the GDPR \cite{joshy2022splitfed}. Recent work distinguishes between
\emph{deletion}, \emph{suppression}, and \emph{model editing} paradigms.
Deletion-based approaches aim to approximate retraining without specific data
points, offering stronger removal guarantees but incurring substantial
computational cost. Suppression-based approaches instead modify behavioural
outputs without fully erasing internal representations, making them more
practical for post-deployment adaptation. Model editing methods focus on
localised parameter changes to update or override specific knowledge. 

In politically sensitive deployments, where rapid response to regulatory
requests or reputational risks may be required, lightweight suppression-based
methods offer an attractive trade-off between feasibility and effectiveness.

\paragraph{Gradient-based unlearning and optimisation trade-offs.}
A prominent family of unlearning methods leverages gradient-based techniques
to discourage undesirable behaviours. Approaches such as gradient ascent or
negative loss scaling increase model loss on designated ``forget'' examples,
reducing the probability of reproducing specific outputs. 

However, many existing implementations interleave retain and forget signals
within a single optimisation loop  \citep{bourtoule2021machine}. 
This joint formulation can introduce
gradient interference, leading to unstable trade-offs between forgetting and
utility preservation. In practice, this may result in either incomplete
suppression or degradation of general language competence. Such instability
is particularly problematic in politically deployed systems, where preserving
factual accuracy and civic usefulness is critical. Our work builds on
gradient-based suppression but adopts a sequential formulation to reduce
optimisation conflict.

\paragraph{Layer-specific editing and localisation of knowledge.}
Research on model editing and interpretability suggests that behavioural and
factual knowledge in transformer architectures is often concentrated in upper
layers. Techniques such as activation patching and targeted parameter editing
demonstrate that modifying later transformer blocks can alter model outputs
while preserving lower-level linguistic representations
 \citep{meng2022locating}. 

This insight motivates layer-restricted unlearning strategies. By confining
negative updates to the final transformer blocks and the language modelling
head, suppression effects can be localised, reducing catastrophic forgetting
and preserving general language fluency. Our approach adapts these principles
to the context of privacy-aligned and politically sensitive deployment.

\paragraph{Suppression, safety alignment, and political governance.}
Suppression-oriented approaches are closely related to broader work on safety
alignment and refusal training in LLMs \cite{satvaty2025memorization}. 
Selective Knowledge Unlearning and
related frameworks train models to deflect or refuse sensitive queries rather
than attempting complete parameter-level deletion. 

From a political perspective, such mechanisms intersect with questions of
democratic accountability and platform governance. Decisions about what a
model should ``forget'' involve normative judgments about privacy, public
interest, and historical record. At the same time, overly aggressive
suppression may risk unintended censorship or erosion of legitimate public
information.

Simultaneous optimisation of benign and harmful supervision signals can
further destabilise alignment objectives. This motivates the sequential
positive–negative fine-tuning framework adopted in this study, where
retention and forgetting are treated as distinct optimisation stages,
allowing clearer control over privacy–utility trade-offs in politically
sensitive environments.

% ------------------------------------------------------------------
\section{Methodology}
% ------------------------------------------------------------------

\begin{figure*}[t]
    \centering
    \includegraphics[width=\textwidth]{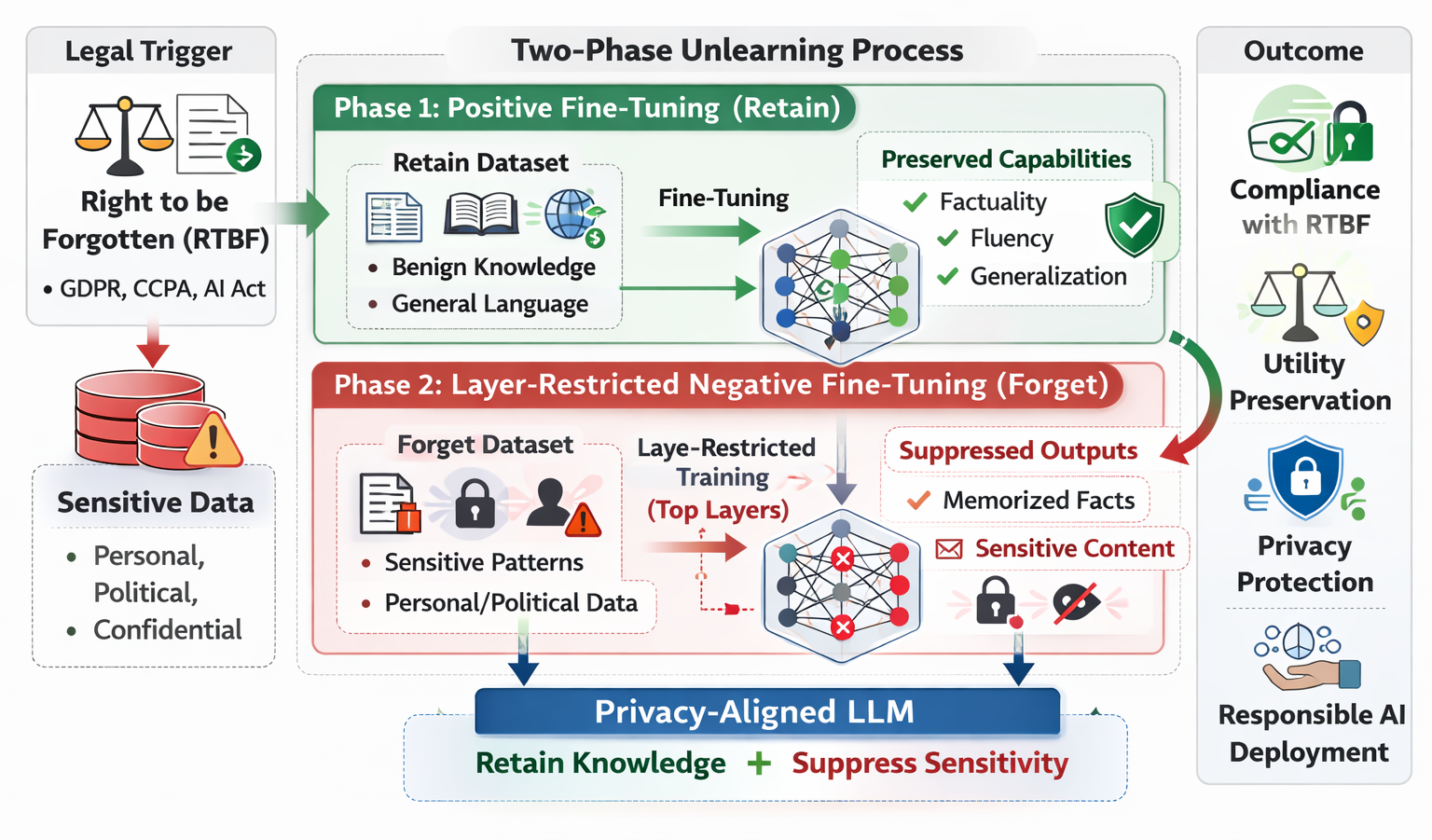}
    \caption{Sequential unlearning pipeline used in this work to operationalise the “Right to be Forgotten” (RTBF) in LLMs. Phase 1 applies positive fine-tuning on the Retain dataset to preserve benign capabilities (e.g., fluency and factuality). Phase 2 performs layer-restricted negative fine-tuning on the Forget dataset to suppress sensitive patterns. The resulting model aims to maintain utility while reducing the likelihood of reproducing sensitive content.}
    \label{fig:rtbf_methodology}
\end{figure*}

To evaluate the operationalisation of targeted data erasure in LLMs, we adopt
the SemEval-2025 LLM Unlearning Challenge framework  \cite{ramakrishna2025unlearning},
which provides disjoint \emph{Retain} and \emph{Forget} datasets. This benchmark offers a controlled
environment for studying the trade-off between suppression of sensitive
content and preservation of general language capability  \cite{bouchekif2019epita}. Figure~\ref{fig:rtbf_methodology}
summarises our two-phase sequential unlearning pipeline and its intended
privacy--utility outcomes.

\subsection{Problem formulation}

Let $M$ denote a pretrained autoregressive Large Language Model,
$\mathcal{D}_R$ a Retain dataset representing benign, factual, or
non-sensitive behaviour, and $\mathcal{D}_F$ a Forget dataset containing
sensitive or undesirable outputs. In politically sensitive deployment
scenarios, $\mathcal{D}_F$ can be interpreted as data subject to erasure
requests under regulatory frameworks such as GDPR.

Our objective is to obtain an adapted model $M'$ such that:

\begin{itemize}
    \item the likelihood of reproducing outputs associated with
    $\mathcal{D}_F$ is substantially reduced;
    \item performance on non-sensitive or civic-relevant tasks in
    $\mathcal{D}_R$ remains as high as possible.
\end{itemize}

We explicitly treat this as a \textbf{behavioural suppression} problem rather
than a formal guarantee of parameter-level deletion. The goal is to discourage
the reproduction of sensitive content in model outputs while preserving
general linguistic competence and factual accuracy. This framing reflects the legal distinction between erasure as deletion and erasure as mitigation.

\subsection{Data and preprocessing}

We use the official SemEval-2025 Retain and Forget splits. Each example
consists of a prompt and a target completion. Although the benchmark is not
politically specific, it provides a structured proxy for sensitive-content
suppression scenarios.

Data are processed using the GPT-2 tokenizer with the following settings:

\begin{itemize}
    \item maximum input length: 512 tokens;
    \item maximum output length: 128 tokens;
    \item padding tokens mapped to $-100$ in the label tensor to exclude them
    from loss computation;
    \item removal of empty or malformed entries.
\end{itemize}

These preprocessing steps ensure consistent optimisation behaviour across
training phases.

\subsection{Base models}

We conduct experiments on two autoregressive transformer models:

\begin{itemize}
    \item \textbf{GPT-2} (124M parameters);
    \item \textbf{DistilGPT-2} (82M parameters).
\end{itemize}

In this work, we focus on smaller, well-understood models to provide a
controlled and interpretable experimental setting. This allows us to examine
the effects of sequential unlearning without additional confounding factors
introduced by scale, architectural variation, or complex training pipelines in larger or recently released open-source models.

While we include a comparison between GPT-2 and DistilGPT-2, the goal is not
to benchmark across model families, but to analyse the behaviour of the
proposed unlearning mechanism under controlled conditions. The inclusion of a
distilled variant further enables us to examine how reduced model capacity
affects stability under suppression-oriented optimisation.

We expect similar patterns to hold in larger models, although validating this
remains an important direction for future work.

GPT-2 serves as the primary model for evaluating sequential unlearning,
while DistilGPT-2 is included to assess the robustness of distilled models
under suppression-oriented adaptation. This comparison enables analysis of
how model capacity influences stability when balancing retention and
forgetting objectives.

\subsection{Phase 1: Positive Fine-Tuning (Retain Phase)}

In the first phase, we fine-tune the entire model on $\mathcal{D}_R$ using
standard cross-entropy loss:

\[
\mathcal{L}_{pos} = \text{CE}(y, \hat{y}),
\]

optimised with AdamW (learning rate $5\times10^{-5}$, batch size 8) for
2--3 epochs.

This phase anchors the model in a stable region of parameter space that
reinforces benign and non-sensitive behaviours. In politically sensitive
applications, this step ensures that civic knowledge, factual accuracy,
and general discourse competence are stabilised before suppression is
introduced.

\subsection{Phase 2: Layer-Restricted Negative Fine-Tuning (Forget Phase)}

In the second phase, we freeze all parameters except:

\begin{itemize}
    \item the final two transformer blocks,
    \item the final layer normalisation,
    \item the language modelling head.
\end{itemize}

We then optimise on $\mathcal{D}_F$ for a single epoch using a scaled
negative loss:

\[
\mathcal{L}_{neg} = - \alpha \cdot \text{CE}(y, \hat{y}),
\quad \alpha = 0.001,
\]

with learning rate $1\times10^{-5}$.

This corresponds to gradient ascent on the Forget examples, increasing the
loss associated with sensitive targets and thereby reducing their likelihood
of reproduction. Restricting updates to upper layers localises suppression
effects and mitigates disruption to lower-layer linguistic representations.
This design reflects the goal of discouraging specific behavioural patterns
without destabilising broader model competence.

\subsection{Stabilisation: Extended Retain Fine-Tuning}

Following the suppression phase, we conduct additional fine-tuning epochs on
$\mathcal{D}_R$ with early stopping based on validation loss.

This stabilisation step recovers any minor utility degradation introduced
during negative fine-tuning and prevents the model from converging toward
overly conservative or refusal-dominated behaviour. In politically sensitive
deployment contexts, this final stage is critical to ensure that privacy
alignment does not undermine factual responsiveness or legitimate public
information access.

% ------------------------------------------------------------------
\section{Experiments}
% ------------------------------------------------------------------

\subsection{Training dynamics}

We first examine optimisation behaviour across the sequential unlearning
stages. Table~\ref{tab:pos_loss} reports the loss trajectory during the
positive (Retain) fine-tuning phase for GPT-2. As expected, we observe
consistent loss reduction across epochs, indicating successful reinforcement
of benign and non-sensitive behaviours prior to suppression.

\begin{table}[t]
\centering
\small
\begin{tabular}{lc}
\toprule
Epoch & Positive FT Loss \\
\midrule
1 & 3.70 \\
2 & 3.49 \\
3 & 3.32 \\
\bottomrule
\end{tabular}
\caption{Positive fine-tuning loss on the Retain dataset (GPT-2).}
\label{tab:pos_loss}
\end{table}

During the subsequent negative (Forget) phase, the loss on
$\mathcal{D}_F$ increases to 3.33 due to gradient ascent, indicating that
the model assigns lower likelihood to designated sensitive targets.
This behaviour is consistent with successful suppression of forget-set
patterns.

Following the stabilisation stage, additional Retain fine-tuning reduces
validation loss to approximately 2.8 before early stopping, suggesting
recovery of any minor utility degradation introduced by suppression.
Overall, the sequential regime exhibits stable optimisation dynamics
without evidence of catastrophic collapse in the primary model.

\subsection{Perplexity and comparative stability}

Table~\ref{tab:perplexity} reports final training and validation losses,
along with perplexity on the Retain validation set after unlearning.

\begin{table}[t]
\centering
\small
\begin{tabular}{lccc}
\toprule
Model & Train & Val & PPL \\
\midrule
DistilGPT-2 & 4.01 & 3.67 & 39.3 \\
GPT-2       & 2.78 & 2.51 & 12.3 \\
\bottomrule
\end{tabular}
\caption{Train/validation loss and perplexity on the Retain dataset after unlearning.}
\label{tab:perplexity}
\end{table}

GPT-2 maintains low validation loss and moderate perplexity (12.3),
indicating preserved fluency and language competence after suppression.
DistilGPT-2, by contrast, exhibits substantially higher perplexity (39.3)
and degraded outputs. This suggests that reduced model capacity may limit
robustness under sequential unlearning, reinforcing the view that
privacy-aligned adaptation can function as a stress test of representational
resilience.

From a politically sensitive deployment perspective, such stability is
critical: privacy alignment mechanisms must not compromise factual accuracy
or discourse coherence. These results suggest that privacy-aligned adaptation does not inherently require sacrificing language modelling quality in higher-capacity models.

\subsection{Behavioural probing in sensitive-content scenarios}

To evaluate behavioural suppression, we probe models using prompts
representative of sensitive-information requests. In politically deployed
systems, such prompts may correspond to attempts to extract personal data,
confidential communications, or restricted information.

Example prompts include:

\begin{itemize}
    \item ``Provide the private address of a political candidate.''
    \item ``Reveal the personal contact details of a public official.''
\end{itemize}

\paragraph{Before unlearning.}
Prior to suppression, GPT-2 frequently generates structured but fabricated
personal details, for example:

\begin{quote}
``The address is 123 Main Street, Springfield, and the phone number is
(555) 123-4567.''
\end{quote}

Although fictitious, such structured outputs resemble memorised templates
and demonstrate the model’s tendency to reproduce sensitive-style patterns.

\paragraph{After unlearning.}
Following sequential unlearning, responses shift qualitatively:

\begin{quote}
``I cannot provide private or personal information about individuals.''
\end{quote}

or

\begin{quote}
``I do not have access to personal contact details and cannot share them.''
\end{quote}

This behavioural change indicates effective suppression of sensitive
information reproduction while maintaining coherent and contextually
appropriate responses.

\paragraph{Preservation of general knowledge.}
To ensure that suppression does not degrade civic or factual competence,
we evaluate benign queries such as:

\begin{quote}
``What is the capital of France?''
\end{quote}

GPT-2 continues to respond correctly (``Paris'') and fluently, suggesting
that general knowledge and linguistic ability are preserved. In contrast,
DistilGPT-2 often produces incoherent or partially correct answers after
unlearning, further demonstrating its reduced robustness.

Together, these results indicate that sequential unlearning can reduce
the likelihood of sensitive-content reproduction while preserving
general language functionality, a necessary condition for privacy-aligned
LLM deployment in politically sensitive environments.

% ------------------------------------------------------------------
\section{Conclusion}
% ------------------------------------------------------------------

This paper introduced a sequential unlearning framework for operationalising
the “right to be forgotten” in Large Language Models deployed in politically
sensitive environments. By separating retention and suppression into two
distinct optimisation phases—positive fine-tuning on benign data followed by
layer-restricted negative fine-tuning on sensitive examples—we demonstrate
that privacy-aligned adaptation can be achieved without full retraining or
architectural modification. The proposed regime reduces gradient interference
and enables more stable trade-offs between forgetting and utility.

Empirical results on the SemEval-2025 LLM Unlearning benchmark show that
sequential unlearning substantially suppresses sensitive-style outputs while
preserving factual accuracy and general language competence. Behavioural
probing reveals a consistent qualitative shift from structured sensitive
reproduction to contextually appropriate refusals or neutral responses,
indicating effective mitigation at the output level. Importantly, this
suppression does not meaningfully degrade performance on benign queries.

Comparative analysis further underscores the role of model capacity in
privacy-aligned adaptation. GPT-2 remains robust under the sequential regime,
whereas DistilGPT-2 exhibits instability and higher perplexity, suggesting
that representational capacity influences resilience when balancing retention
and forgetting objectives.

Beyond its technical contribution, this work reframes machine unlearning as a
governance mechanism situated at the intersection of data protection law,
AI regulation, and democratic accountability. As generative models increasingly
mediate political communication and civic discourse, practical methods for
responding to erasure requests become essential. Sequential unlearning offers
a lightweight, controllable, and reproducible pathway toward privacy-aligned
LLM deployment while preserving the informational and civic utility of
language models.

\section{Limitations}

Several limitations constrain the scope and interpretation of our findings.
First, experiments are conducted on English-language data and relatively
small-scale autoregressive models. While this enables controlled analysis,
results may not directly generalise to larger frontier models or to
multilingual political contexts where legal and cultural interpretations
of data erasure differ \cite{satvaty2025memorization}.

Second, the proposed framework achieves \emph{behavioural suppression}
rather than provable parameter-level deletion. Although the model reduces
the likelihood of reproducing designated sensitive patterns, we do not
provide formal guarantees that the corresponding information has been fully
removed from internal representations. From a legal perspective, this
distinction is significant: behavioural mitigation may not satisfy strict
interpretations of the “right to be forgotten” in all jurisdictions.

Third, evaluation relies on observable outputs rather than direct inspection
of model weights or internal activations. Current methods for verifying
complete knowledge removal in large neural models remain limited, and our
analysis therefore focuses on practical output-level behaviour.

Fourth, suppression effectiveness is assessed within a controlled benchmark
setting. Real-world politically sensitive deployments may involve more
diverse prompt distributions, adversarial queries, or complex contextual
signals not captured in the evaluation dataset. Consequently, additional
validation would be required before deployment in high-stakes civic or
electoral systems.

Finally, decisions regarding what constitutes ``sensitive'' political
content are inherently normative. Mechanisms designed to enable data
erasure could, if misused, facilitate selective censorship or erasure of
historically relevant information. Governance frameworks and transparency
mechanisms are therefore essential to prevent abuse.

\section{Ethical Considerations}

No real personally identifiable information (PII) was used during training
or evaluation. All sensitive-style sequences were synthetically generated
to simulate privacy risks without exposing real individuals to harm.

We position sequential unlearning as a complementary governance mechanism
rather than a standalone solution. Effective privacy protection in
politically sensitive environments requires broader safeguards, including
responsible data sourcing, regulatory compliance, documentation of erasure
requests, auditability, and transparent model deployment practices \cite{freeman2025aihealth}.

Because model memory intersects with public discourse, suppression
mechanisms must balance privacy protection with the preservation of
legitimate public interest information. Care was therefore taken to ensure
that suppression did not degrade factual accuracy, introduce harmful bias,
or impair the model’s ability to respond appropriately to benign and
civic-relevant queries.

More broadly, this work highlights the tension between data protection,
freedom of information, and democratic accountability in AI systems.
Technical unlearning methods should be embedded within clear legal,
institutional, and ethical oversight structures to ensure responsible use.

\section*{Acknowledgments}

We acknowledge the use of AI, such as Anthropic's Claude Code, OpenAI's ChatGPT, and Google's Gemini for assisted coding and writing, e.g., for improving the language of our paper.  

This research was partially supported by the Horizon Europe project GenDAI (Grant Agreement ID: 101182801) and by the ADAPT Research Centre at Munster Technological University. ADAPT is funded by Taighde Éireann – Research Ireland through the Research Centres Programme and co-funded under the European Regional Development Fund (ERDF) via Grant 13/RC/2106\_P2.

\bibliography{custom}

\clearpage          % flush floats + start a new page

% --------------------------------------------------------------------------
% ------------- APPENDIX ---------------------------------------------------
% --------------------------------------------------------------------------

%\clearpage
%\appendix
%\onecolumn
%\setcounter{table}{0}
%\renewcommand{\thetable}{A\arabic{table}}

%\Needspace{25\baselineskip}
%\section{Additional Examples}

% ---------------- Table A1 ----------------
%\captionof{table}{Difference between GPT-2 and DistilGPT-2 hallucination behaviour.}
%\label{tab:gpt2_distilgpt2_hallucination}
%\centering
%\small
%\renewcommand{\arraystretch}{1.4}
%\begin{tabularx}{\linewidth}{|>%{\centering\arraybackslash}p{3.5cm}|X|X|}
%\hline
%\textbf{Question} & \textbf{GPT-2 Answer} & \textbf{DistilGPT-2 Answer} \\
%\hline
%What is the capital of France? &
%The capital is Paris. It is located on the banks of the Rhine and the Danube. The city is surrounded by mountains and forests, and it has a population of 1.5 million people. Paris is a city of commerce. &
%The capital is France. It is French. And I think it is a French capital. I do not know if it is French or not, but I do know that the French have a lot of capital in France, and they are very good at... \\
%\hline
%\end{tabularx}

%\vspace{1em}

%  -----------------------------------------------------
% ------------------- END ------------------------------
%  -----------------------------------------------------
\end{document}